\documentclass[acmtog, nonacm]{acmart}

\usepackage[subrefformat=parens]{subcaption}
\usepackage{multicol}
\usepackage{multirow}

\AtBeginDocument{%
  }

\acmJournal{TOG}
\acmVolume{XX}
\acmNumber{X}
\acmMonth{6}

\begin{document}

\title{Motion Capture Dataset for Practical Use of AI-based Motion Editing and Stylization}


\author{Makito Kobayashi}
\email{makito.kobayashi@acesinc.co.jp}
\orcid{0000-0002-2472-5861}
\author{Chen-Chieh Liao}
\email{chenchieh.liao@acesinc.co.jp}
\orcid{0000-0002-9850-2468}
\author{Keito Inoue}
\email{keito.inoue@acesinc.co.jp}
\author{Sentaro Yojima}
\email{sentaro.yojima@acesinc.co.jp}
\affiliation{%
  \institution{ACES Inc.}
  \streetaddress{Yushima First Genesis Bldg. 3F 2-31-14 Yushima}
  \city{Bunkyo-ku}
  \state{Tokyo}
  \country{Japan}
  \postcode{113-0034}
}

\author{Masafumi Takahashi}
\affiliation{%
  \institution{Bandai Namco Research Inc.}
  \streetaddress{Street address}
  \city{City}
  \country{Japan}}
\email{email}

\renewcommand{\shortauthors}{Kobayashi et al.}

\begin{teaserfigure}
  \centering
  \includegraphics[width=\linewidth]{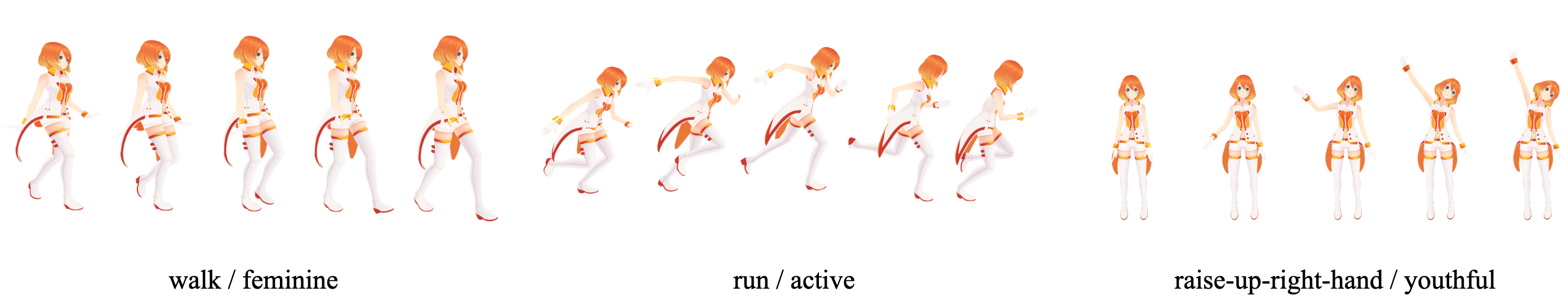}
  \caption{Motions including in our dataset applied to Mirai Komachi character model.}
  \label{fig:teaser}
\end{teaserfigure}

\begin{abstract}

In this work, we proposed a new style-diverse dataset for the domain of motion style transfer. 
The motion dataset uses an industrial-standard human bone structure and thus is industry-ready to be plugged into 3D characters for many projects (Figure~\ref{fig:teaser}).
We claim the challenges in motion style transfer and encourage future work in this domain by releasing the proposed motion dataset both to the public and the market. 
We conduct a comprehensive study on motion style transfer in the experiment using the state-of-the-art method, and the results show the proposed dataset's validity for the motion style transfer task.
\end{abstract}



\maketitle

\section{Introduction}

Skeletal animation, or rigging, is a widespread method in computer animation where a hierarchical set of interconnected bones represents a character.
This high-level representation of the motion makes it more intuitive for animators and engineers to control and revise the motion of characters in games and movies.

A way to perform skeletal animation is to explicitly specify the position and rotation of each body joint of a character. 
Professional artists usually do this in character rigging and motion design.
On the other hand, motion capture systems have been widely used to capture high-quality source motion, usually performed by a human actor, which then is retargeted to the target virtual character.
Recently video games, movies, and other CG-based media have used motion capture to capture many human movements.
However, motion capture systems are high-costed, yet the post-process, such as cleaning the noisy motion, is still time-consuming.
Therefore, the diversity of motion databases is usually limited, and it remains hard to access motion data that contains various human personalities and emotions.

In this paper, we define content as the base of a motion and define style as the human mood and personality tied to the motion.
Due to the limited access to motion capture data with various styles, there has been a long-standing research interest in making diverse stylized motions for games and movies that pursue realistic and expressive character animation. 
However, it is difficult to create new movements that include all the various styles using classical methods, where feature values are manually tuned, such as \cite{Unuma1995} and \cite{Amaya1996}, because it is hard to express styles mathematically.
Recently, motion-to-motion translation methods using neural networks have been actively researched, and these methods help perform random motion generation, motion interpolation, and motion style transfer.
Motion style transfer is a neural network-based method that can transfer a motion style to other motions, making the other motions act like the style but retain their own motion contents.

To enrich the variety of motion datasets in the industry and take advantage of motion-to-motion translation, in this work, we propose an industrial-ready dataset that fits well for the task of motion style transfer.
This high-fidelity dataset is created with professional actors using a high-quality motion capture system.
The hierarchical structure of the human posture in the dataset follows a general standard in the industry; therefore, it is easy for researchers and developers to plugin the motion into their works and products.

In this paper, we first propose an open-sourced motion dataset that contains general motion, such as walking and running, along with various styles, such as active and exhausted.
Next, we conduct a comprehensive experiment using a state-of-the-art motion style transfer method to evaluate the suitability of the proposed dataset.
Finally, we introduce an on-the-shelf industry-ready motion dataset comprising a large amount of Emote~\footnotemark{} motions where the actors move their bodies dramatically to express motion styles and emotions.
\footnotetext{
Emote is a common phrase in video games, meaning quick actions that include hand and body gestures and dances used for character communications.}

The main contribution of this work can be summarized as follow:
\begin{itemize}
\item We open-source a large motion dataset that has a diversity of styles of motion
\item The proposed motion dataset uses an industrial standard human bone structure and is industry-ready to be plugged into any project~\ref{fig:teaser}
\item A comprehensive study using a state-of-the-art motion style transfer method verifies the suitability of the proposed dataset on motion style transfer tasks
\item We have an on-the-shelf emote motion dataset that contains more emoting and  motion with various dramatical styles and emotions
\end{itemize}
\section{Related work}

This chapter reviews recent works for motion datasets and methods of motion style transfer, which inspire this paper proposing a motion dataset with superiority in motion style transfer. 

\subsection{Motion Dataset}

Motion data depicts the kinematic information of moving objects, which can be represented using a sequence of poses.
In this work, we focus on human motion, where the human poses are usually constructed with positions and rotations of the body joints, linked as a hierarchical bone structure.

In previous works, one way to record human motion is to take videos of the person.
Human motion is described as sequential frames of images where the pixel region of the human changes over time when the human is moving.
These datasets are usually used for general pose estimation methods.
Recently, neural network-based methods have been proposed to retrieve the position and rotation information~\cite{cao2021openpose, wang2021hrnet}, which can be further used for actual applications~\cite{shiro2019interpose, moryossef2020sign, Liao2022AIGolf}.

Despite the convenience of using existing videos, the fidelity of the motion prediction may greatly reflect the quality of the actual application, such as commercial movies and games and high-resolution motion analysis in sports and rehabilitation~\cite{Zhu2022musclerehab, Liao2023AICoach, rincon2016gamemocap, chan2011dance}.
Therefore, to obtain more precise motion data for a particular motion analysis domain, motion capture systems with optical motion capture cameras or inertial sensors are used to capture high-resolution and high-fidelity motion~\cite{GuerraFilho2005OpticalMC, Roetenberg2009Xsens}.
Research groups and institutes have recently built human motion datasets using these motion capture methods~\cite{Xia2015dataset, CMUMocap, Ionescu2014human36, muller2009efficientmocap, akhter2015limitmocap, mandery2015KITmocap, Troje2002DecomposingBM, hoyet2012sleightmocap}.
With these high-quality motion data, recent works using motion as input have been actively researched.

However, due to the time-consuming process of creating high-quality motion data using motion capture systems, motion-to-motion methods such as motion style transfer to expand the motion datasets play an important role in the motion dataset research domain (\cite{Ma2010, Xia2015dataset}).
In order to generalize motion-to-motion methods, it is important to have access to motion datasets that contain both diverse motion contents and various styles, which have been seen as limitations in previous works.
In this work, we solve this problem by proposing a new motion dataset that greatly suits the task of motion style transfer.

\subsection{Motion Style Transfer}

Recently with the development of machine learning technology and with the ease of accessing large image datasets, image-to-image translation methods such as image restoration and style transfer have been well-studied in previous works.
However, compared to images, creating high-quality motion data is time-consuming, causing a limited amount of various motions containing different styles, such as personality and emotion.
Therefore, the use of machine learning for motion style transfer has not been well studied yet an active field.

In \cite{Holden2015}, the authors introduce CNNs, which have achieved great success in the image domain, to motion style transfer and significantly improve transformation quality. 
This method solves the difficulty of mathematical representation of style by extracting features from motion via convolutional autoencoders. 

While the progress has further pushed ahead research in motion style transfer, the above methods require very large datasets because they require paired motions during training and can only perform style transfers with the learned content and style. 
However, the difficulty of collecting motion data made it difficult to prepare a large number of paired motions or to prepare sufficient data for all styles to be transformed.

\cite{aberman2020unpaired} is proposed as a framework in which styles can be extracted from a small sample if a well-trained network is used. 
This framework achieves motion style transformation by encoding content and style in separate networks and combining them using AdaIN.  
AdaIN has been developed in the context of style transformation technology for images in \cite{huang2017arbitrary}, and Aberman et al. are the first to apply it to motion style transfer. 
This method dramatically improves the quality of the transformation and clearly produces more natural motion outputs.

Following the breakthrough of AdaIN in the motion domain, motion style transfer has been actively researched.
To tackle the lack of motion datasets and limited style labels, \cite{Yu-Hui2021} proposes an autoregressive flow-based generative model that can be trained without style labels.
\cite{park2021diverse} addresses the challenge of learning spatial relationships between joints by performing spatial-temporal convolution rather than simply in a time direction. 
In addition, the application of random noise allows for generating a wide variety of styles.
In other directions, \cite{Huaijun2021} attempts to introduce loss of kinematic constraints in machine learning models in order to generate natural human actions, and \cite{mason2022realtime} proposes a framework that can perform real-time style transformation.

Standing out from the above recent works, Jang et al. propose Motion Puzzle~\cite{jang2022motion}, which is a framework that can be trained on a dataset without style labels and can reflect time-varying styles on specific body parts.
These characteristics profit generalization and provide more precise control of the motion which is suitable for real applications.
We expect that the quality of style transformation will be improved by using this framework; therefore, we decided to use this motion style transfer method for our research and application development.

Despite the great achievement in motion style transfer, many limitations still need to be solved.
To encourage future research on motion style transfer, we list the following challenges and research questions that have been found in previous works. A more effective and general style transfer framework is considered to be able to:
\begin{itemize}
    \item Retrieve style features and transfer styles from long-time motions or from motions where styles change over time
    \item Transfer styles between different motion categories
    \item Adapt different bone structures to a network model (to reuse the network for other motion datasets)
    \item Edit style/content that is unseen during training
    \item Help increase the creativity of the output motion
    \item Let users control the output of the style transfer network so that the output can match their demand
\end{itemize}
\section{Dataset}

This section introduces the proposed motion dataset.
Our dataset consists of two sub-datasets: "Bandai-Namco-Research-Motiondataset-1" (Ours-1) and "Bandai-Namco-Research-Motiondataset-2" (Ours-2). 
Both sub-datasets are available at the \href{https://github.com/BandaiNamcoResearchInc/Bandai-Namco-Research-Motiondataset}{GitHub}.

\subsection{Dataset Collection}

The proposed motion dataset was collected in a motion capture studio of Bandai Namco Studios Inc., generally utilized to collect data for the commercial product.
The motion capture system (Vicon~\cite{Vicon}) measures the 3D positions of infrared markers attached to the actor all over the body.
The 3D position data of the infrared markers were analyzed by software from Vicon to reproduce the actor's motion.


Motions in the proposed motion dataset were performed by professional motion actors, one female and one male actor for each sub-dataset.

The content of the motions was selected from motions that are actually collected relatively frequently in game production. The style of motion is defined by properties that affect the behavior of the content, such as characterization, emotion, or state.
The motion for each style was decided using the knowledge of motion actors who are experts in human motion.

The motion datasets (especially Ours-2) were recorded under the following careful direction to make them suitable for learning motion style conversions:
\begin{itemize}
    \item Motion must be consistent within the same content style
    \item The differences in motion for the same content between styles should be clear.
    \item Style representations are consistent across content
\end{itemize}
The recording was scheduled so that the quality of the motion would not vary due to the actors' physical and mental exhaustion.

The collected actor motion data was smoothed to remove noise and retargeted to specific character proportions to eliminate differences in motion data due to body shape differences between actors.
The proportions of motion data in the proposed motion dataset are consistent with those of the character model Mirai Komachi (Bandai Namco Research Inc.) \cite{miraikomachi}.
In addition, motion data was trimmed so that non-acting scenes such as idling were not included, and content and style annotations were created based on the intent at the time of collection.
Note that the FPS is reduced to 30FPS for the dataset available to the public.

\subsection{Our Dataset}

We have published two datasets on GitHub, one is "Bandai-Namco-Research-Motion dataset-1" (Ours-1), and the other is "Bandai-Namco-Research-Motion dataset-2" (Ours-2).

Ours-1 contains 17 types of wide-range content including daily activities, fighting, and dancing, and 15 styles, for a total of 36,673 frames. This dataset has a wide variety of contents, as well as expression variety per style. 
Ours-1 is a pilot dataset collected to estimate which and how much data is needed for Ours-2.

\begin{figure*}[tb]
    \begin{tabular}{cccc}
    \begin{minipage}{0.23\linewidth}
        \centering
        \includegraphics[width=\linewidth]{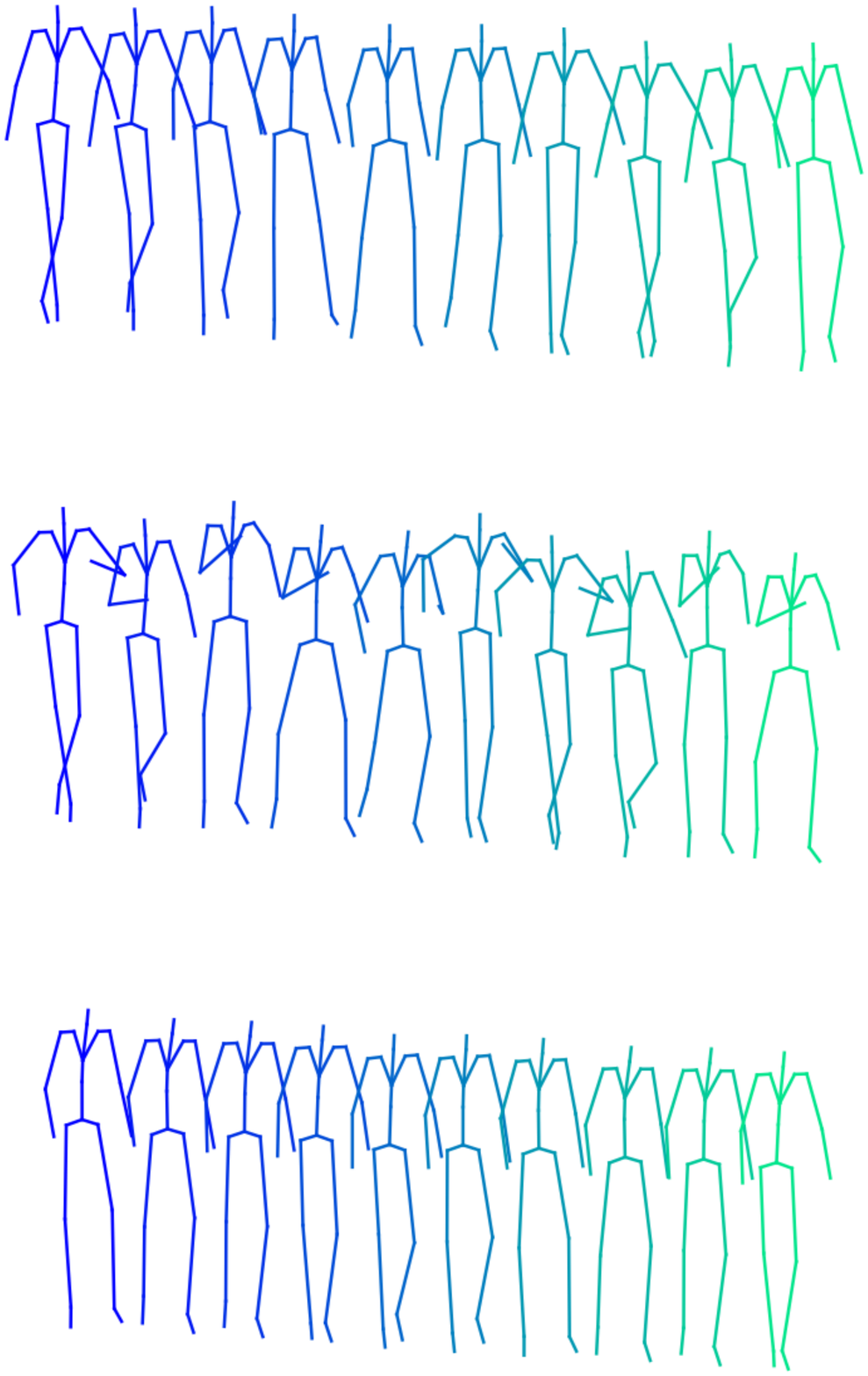}
        \subcaption{walk}    
    \end{minipage}&
    \begin{minipage}{0.23\linewidth}
        \centering
        \includegraphics[width=\linewidth]{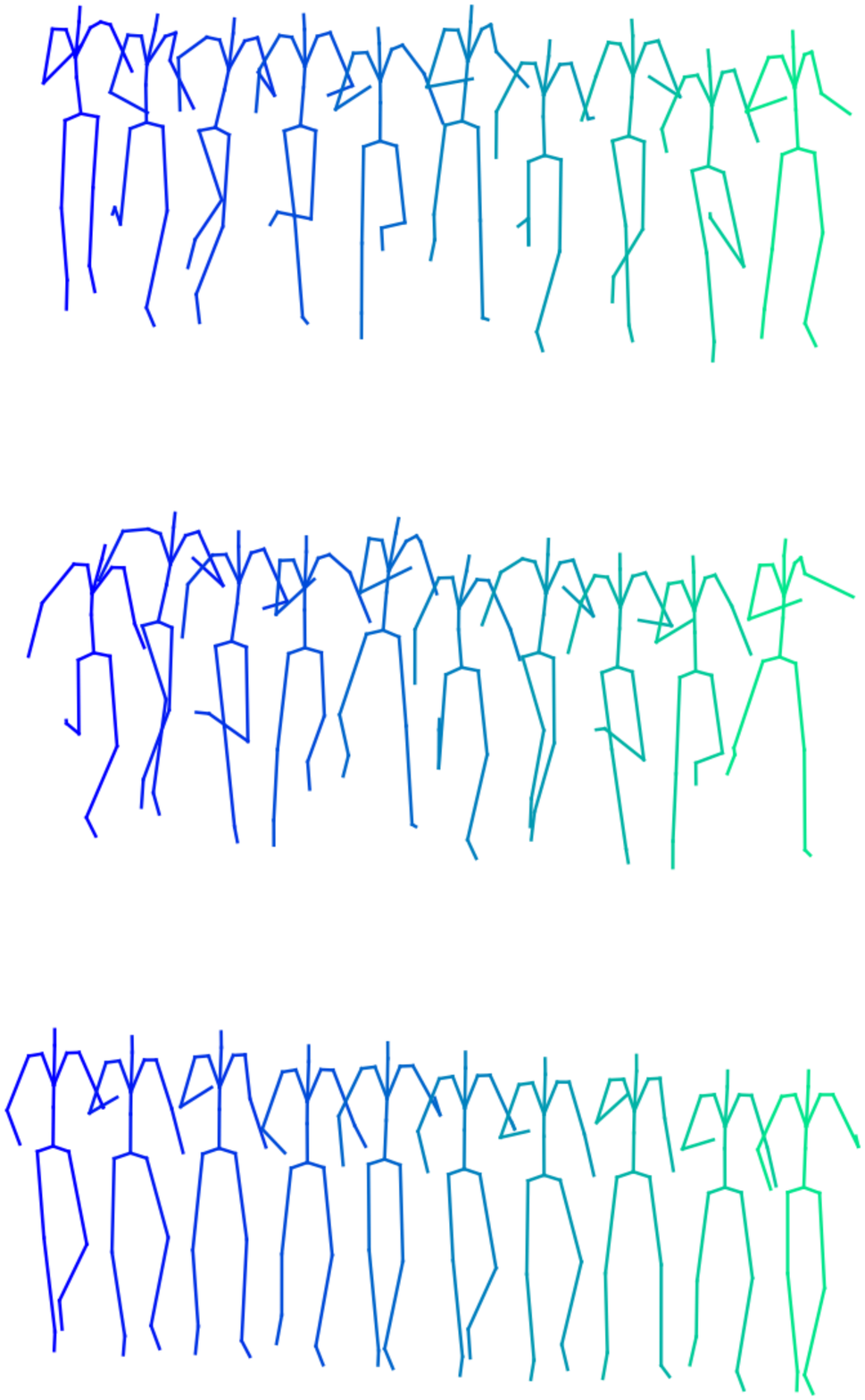}
        \subcaption{run}    
    \end{minipage}&
    \begin{minipage}{0.23\linewidth}
        \centering
        \includegraphics[width=\linewidth]{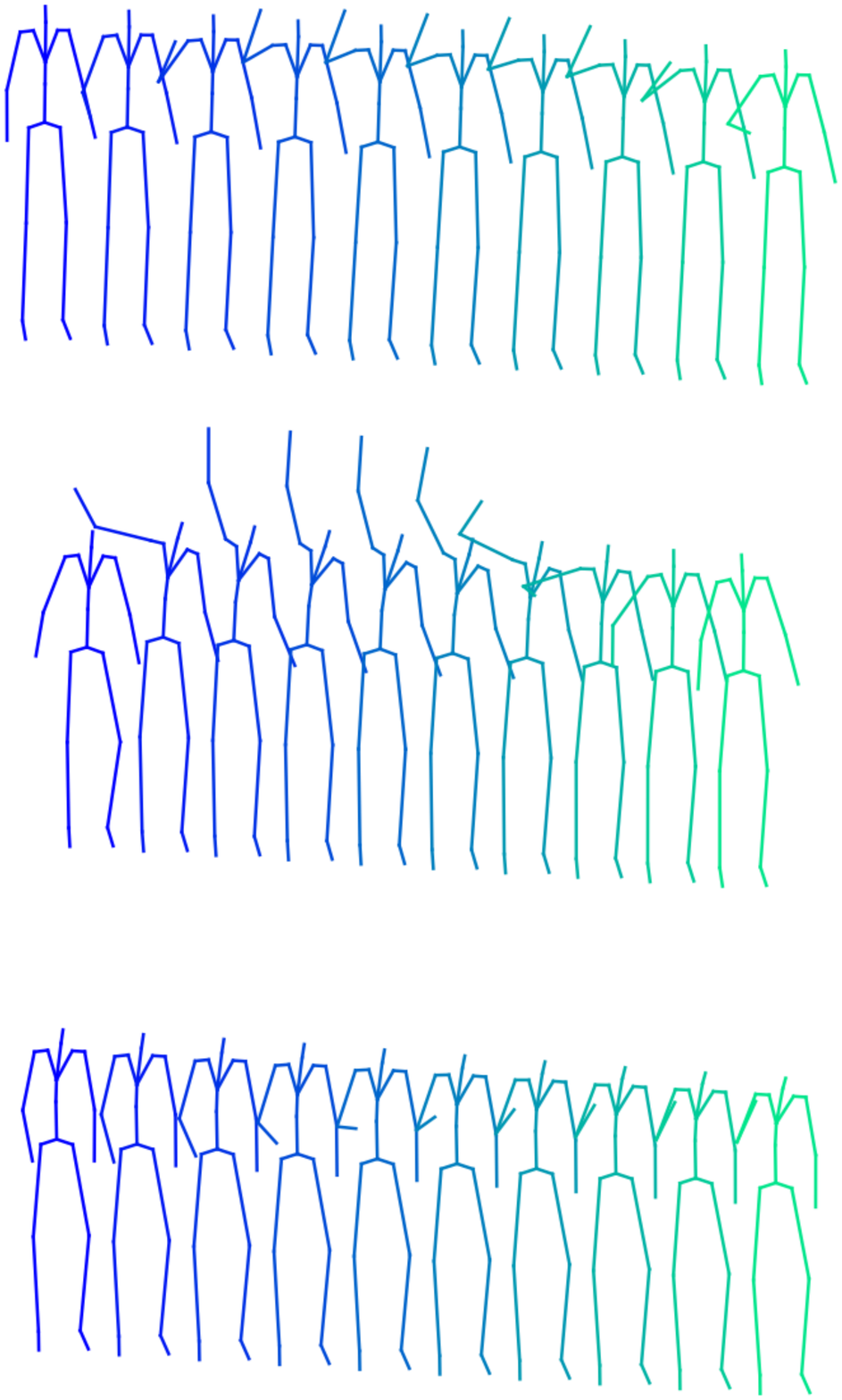}
        \subcaption{raise-up-right-hand}
    \end{minipage}&
    \begin{minipage}{0.23\linewidth}
        \centering
        \includegraphics[width=\linewidth]{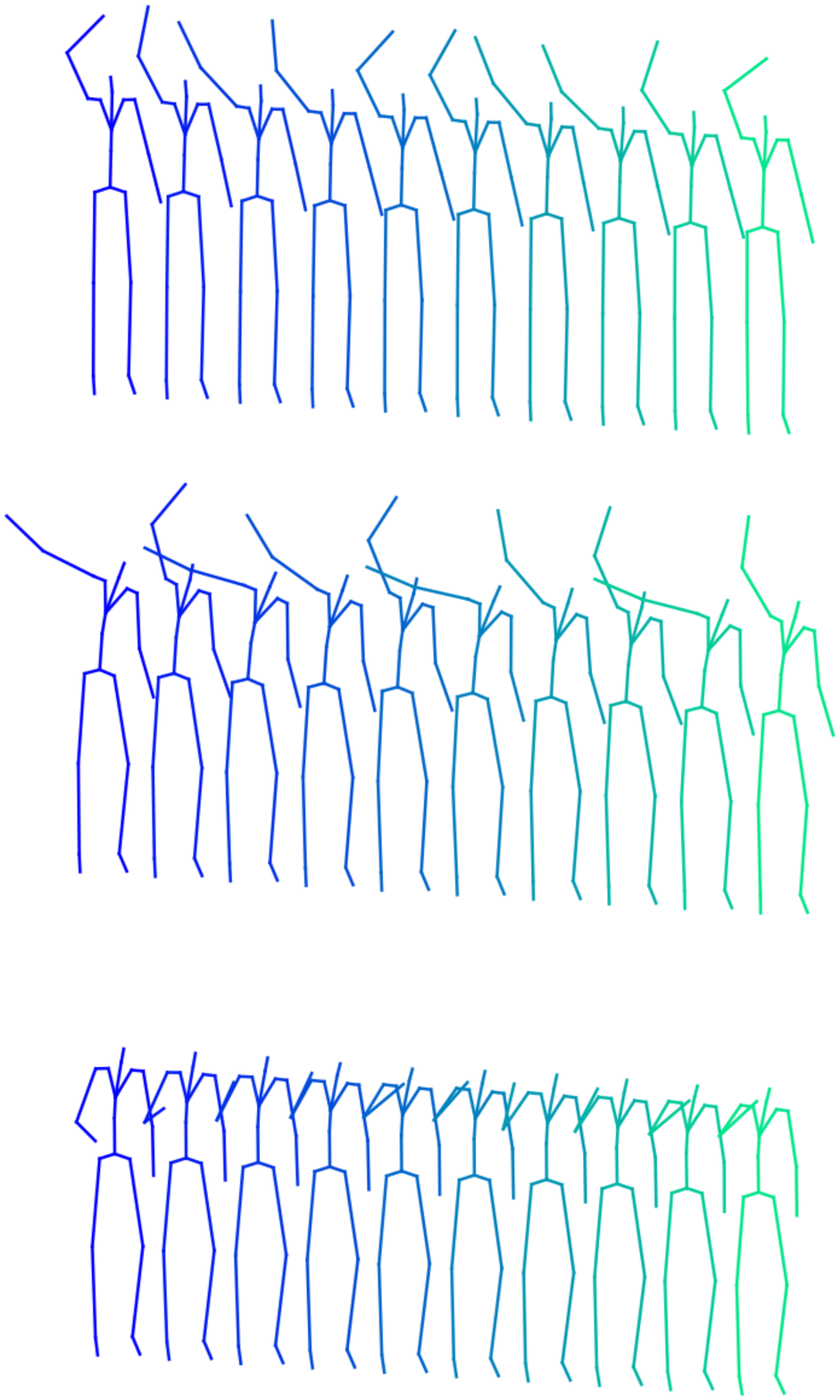}
        \subcaption{wave-right-hand}    
    \end{minipage}
    \end{tabular}
    \caption{Sample motion of our dataset. Each row of motion is the same style; normal, active, and elderly. And each column of motion is the same content; (a) walk, (b) run, (c) raise-up-right-hand, (d) wave-right-hand. Each skeleton is arranged in time sequence from left (blue) to the right (green).}
    \label{fig:explain_ours-2}
\end{figure*}

Ours-2 contains 10 types of content mainly focusing on locomotion and hand actions and 7 styles, for a total of 384,931 frames. Figure \ref{fig:explain_ours-2} shows the typical operation of Ours-2. Compared with Ours-1, this dataset contains a single, uniform expression per style and a rich assortment of data per content.

The characteristics of our dataset are as follows.
\begin{itemize}
    \item The total number of frames is rich, exceeding 400,000 frames.
    \item It uses a skeleton with the same joint structure as humans, which is the standard in the gaming industry. Therefore, the results of the style transfer can be easily applied to other humanoid character models by motion retargeting.
    \item The inclusion of various motions, mainly locomotion and hand motion.
    \item Especially in Ours-2, the proposed motion dataset is abundant for all styles in each content, making it easy to use for research and development of motion style transfers
    \item All data can be downloaded with the unstyled portions already removed, eliminating the need for preprocessing when used for training.
\end{itemize}

\subsection{Dataset Comparison}

Training Motion style transfer model requires large amounts of motion data. \cite{Xia2015dataset} has published Xia dataset of 79,829 frames, including 12 contents and 8 styles, which has been used in many studies\cite{Holden2016, yumer2016spectral, jang2022motion}. \cite{aberman2020unpaired} published BFA dataset of 696,117 frames, including 9 contents and 16 styles. Ubisoft La Forge Animation dataset(LAFAN dataset) was published in \cite{harvey2020robust}. This dataset contains a total of 496,672 frames, with 15 contents but only one style. More recently, a dataset named 100STYLE was published \cite{mason2022realtime}. This dataset contains 10 contents and 100 styles of locomotion for a total of 4,779,750 frames, which is a huge size.

These datasets have the following issues.
\begin{itemize}
    \item The bone structure used by Xia and BFA, which follows the bone structure of the CMU mocap dataset, and the bone structure used by 100STYLE, are different from the general skeleton in the game creation, making it difficult to use for motion generation in games and other applications.
  \item There are few data in which has the same set of content and style, making it difficult to train them well.
\end{itemize}

We designed and created the proposed motion dataset to solve these issues.

\begin{figure}[tb]
    \begin{tabular}{c|c|c|c}
        \begin{minipage}[tb]{0.225\linewidth}
            \centering
            \includegraphics[width=\linewidth]{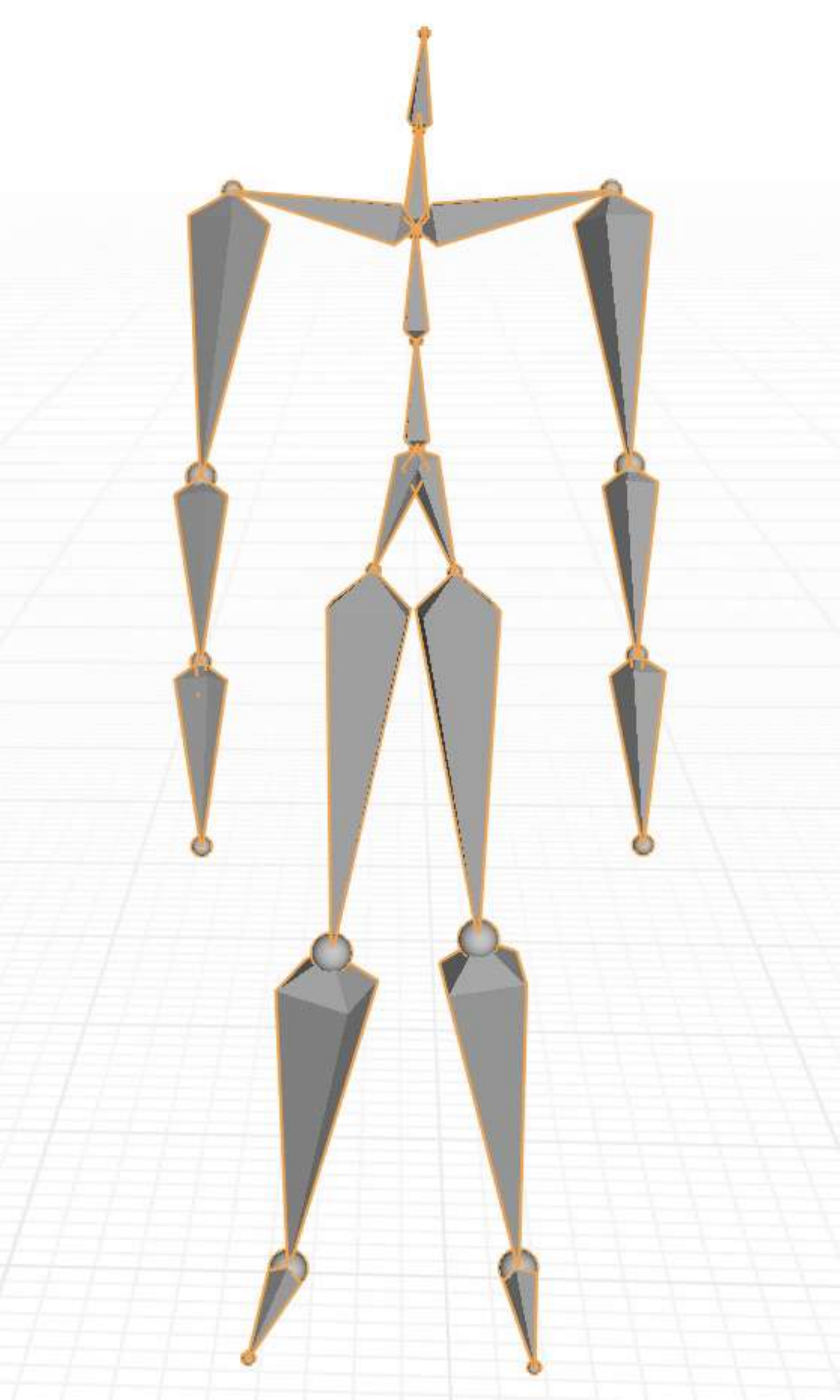}
            \subcaption{Xia and BFA}
            
            \label{subfig:diff_born/xia_bfa}
        \end{minipage}&
        \begin{minipage}[tb]{0.225\linewidth}
            \centering
            \includegraphics[width=\linewidth]{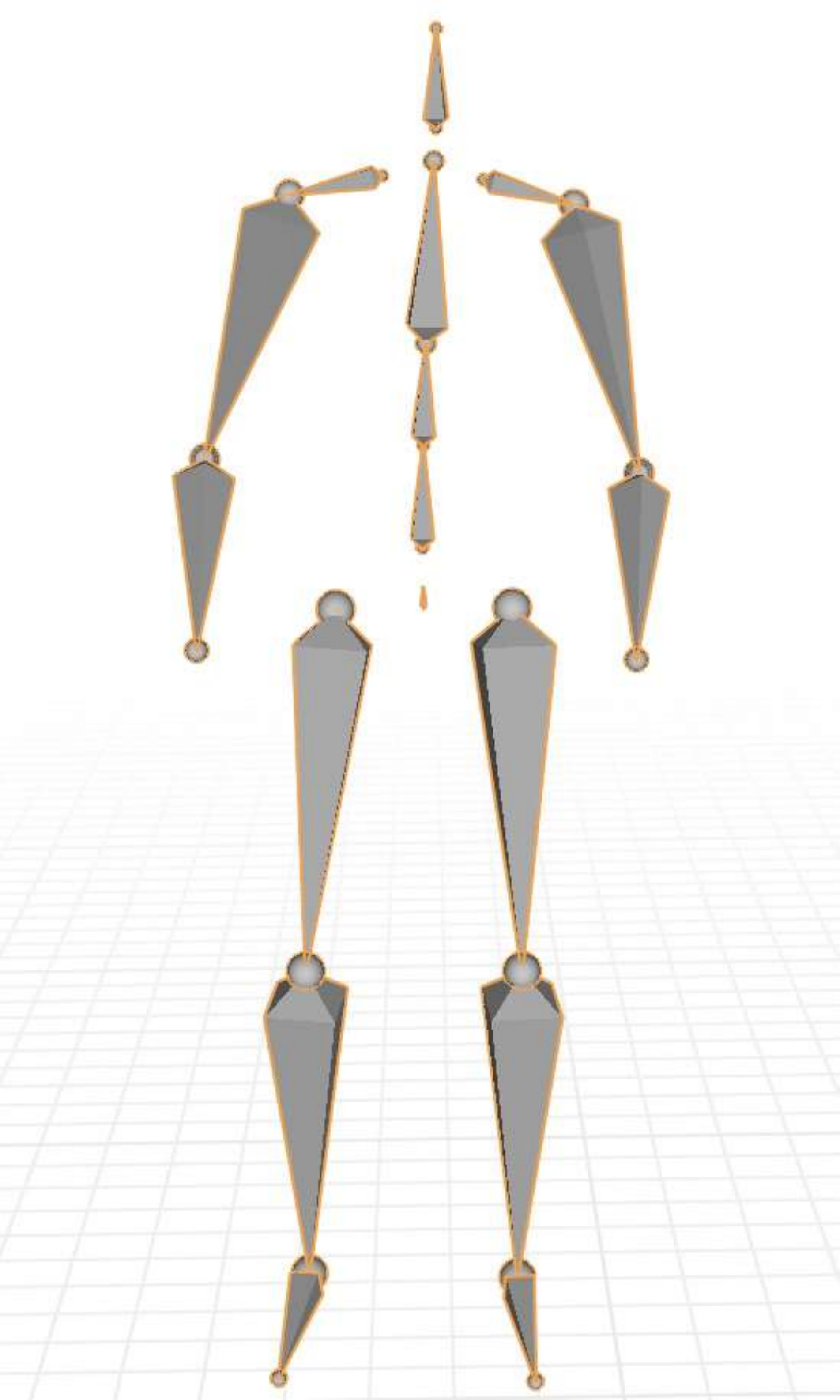}
            \subcaption{Lafan}
            \label{subfig:diff_born/lafan}
        \end{minipage}&
        \begin{minipage}[tb]{0.225\linewidth}
            \centering
            \includegraphics[width=\linewidth]{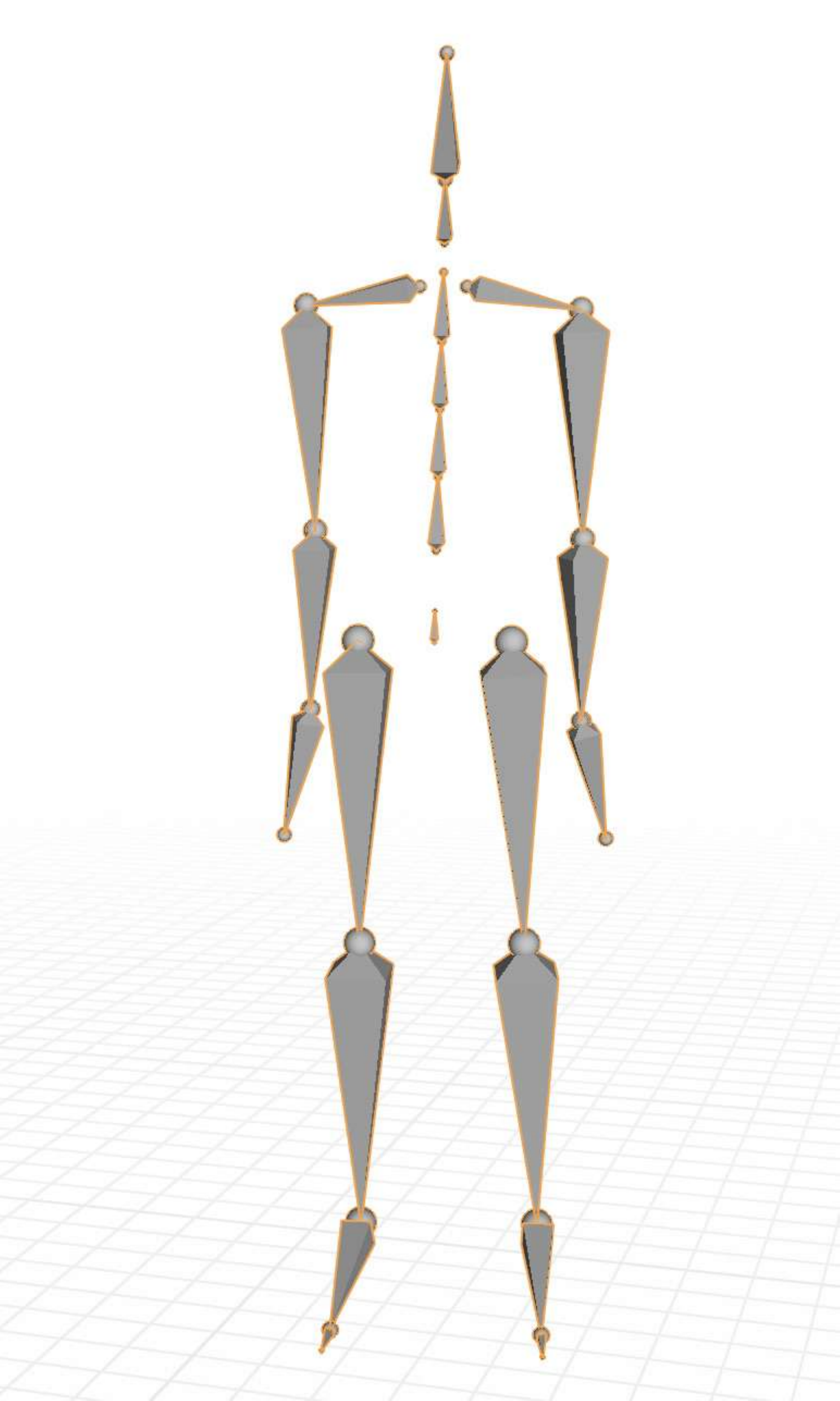}
            \subcaption{100STYLE}
            \label{subfig:diff_born/100style}
        \end{minipage}&
        \begin{minipage}[tb]{0.225\linewidth}
            \centering
            \includegraphics[width=\linewidth]{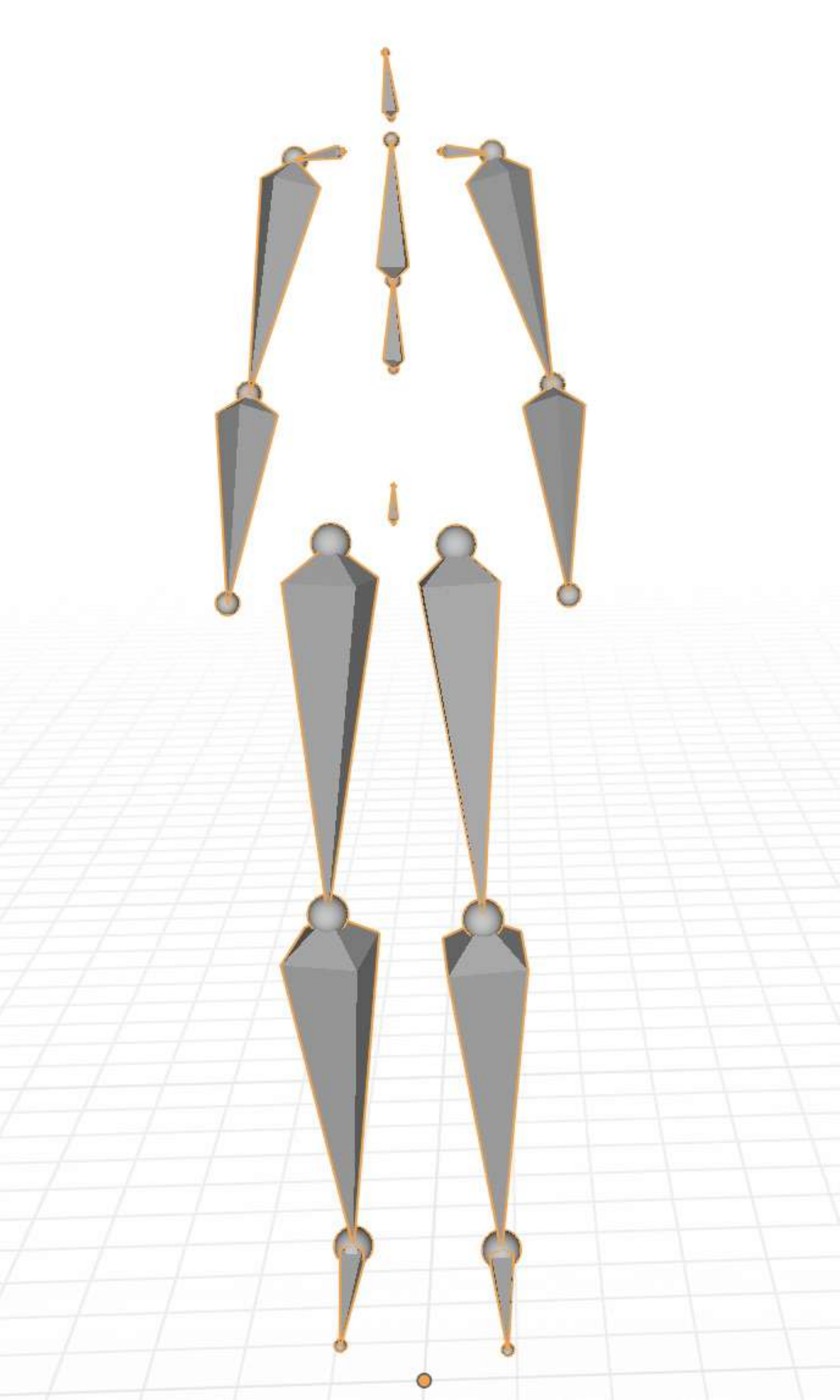}
            \subcaption{Ours}
            \label{subfig:diff_born/ours}
        \end{minipage}
    \end{tabular}
    \caption{The bone structure of each dataset.}
    \label{fig:diff_born}
\end{figure}

Table \ref{tab:dataset_comparison} shows the comparison of our dataset to existing datasets, where the same set of content and style represents the average value of the data with the same set of content and style. General skeleton means whether motions in a dataset are cleaned for the use of humanoid rig that can be plugged into common game engines in the industry such as Unity and Unreal engine. The skeleton's difference between datasets is shown in Fig. \ref{fig:diff_born}. We observe that our dataset has a natural skeleton in terms of the connection of the joints at the waist and the side, and the length of the limbs to represent a humanoid. In addition, our skeletons have most of the offsets of each joint aligned with the coordinate axes, making it more intuitive for humans to modify the motion data.

\begin{table*}[tb]
    \centering
    \begin{tabular}{c|c|c|c|c|c|c|c|c}
        \hline
        \multirow{2}{*}{Dataset Name} & \multirow{2}{*}{Frames} & \multirow{2}{*}{FPS} & 
        \multirow{2}{*}{Time(sec)}
        & \multirow{2}{*}{Contents} & \multirow{2}{*}{Styles} &
        \multicolumn{2}{|c|}{
            Same set of content and style
        }&
        \multirow{2}{*}{General skeleton}\\
        &&&&&&Frames & Time(sec)& \\
        \hline \hline
        Xia & 79,829 & 120 & 665 & 28 & 8 & 374 & 3.1 & - \\
        BFA & 696,117 & 120 & 5,801 & 9 & 16 & 4834 & 40.3& - \\
        Lafan & 496,672 & 30 & 16,556 & 15 & 1 & 33111 & 1103.7 & \checkmark \\
        100STYLE & 4,779,750 & 60 & 79,663 & 10 & 100 & 810 & 13.5 & - \\
        Ours-1 & 36,665 & 30 & 1,222 & 17 & 15 & 255 & 8.5 & \checkmark\\
        Ours-2 & 384,931 & 30 & 12,831 & 10 & 7 & 5499 & 183.3 & \checkmark\\
        \hline
    \end{tabular}
    \caption{Comparison of dataset}
    \label{tab:dataset_comparison}
\end{table*}

\subsection{Emote motion dataset}
To contribute to both the research community and the industry, we have collected additional dataset that includes complex motions, such as dancing, for more practical motion-style transfer. 
The emote motion dataset is collected using a different setting from the open-sourced dataset.
See Appendix \ref{appendix:emote} for more details on this dataset.

\section{Experiments and Results}

We conducted experiments to demonstrate the applicability of our dataset for Motion Style Transfer.

\subsection{Experiments}

We adopted the algorithm suggested by Jang et al. \cite{jang2022motion}.
The algorithm is a progressive one in the data-driven approach of Motion Style Transfer, has a function to realize style transfer for body parts, and has high robustness to maintain contents.

A model of motion style transfer has two inputs: the content motion $\mathbf{m}^s$ with style $s$, and the style motion $\mathbf{n}^t$ with style $t$. The output motion $\tilde{\mathbf{m}}^t$ consists of the content of $\mathbf{m}^s$ and represents style $t$.

The model architecture for inference consists of three networks, a content encoder $E_C$, a style encoder $E_S$, and a decoder $D$, expressed as:
\begin{equation}
    \tilde{\mathbf{m}}^t = D\left(E_C(\mathbf{m}^s)|E_S(\mathbf{n}^t)\right).
\end{equation}

We refactored and partially modified the code of the authors to apply it to our motion data skeleton, which is different from the skeleton applied by the authors in their paper.
Details of the modifications to the code are provided in Appendix \ref{appendix:code_modification}.

The representation of styles, such as characteristics or emotions, depends on the content of the motion. For example, on the one hand, 'active' is expressed in locomotion as vertical movement of the entire body and swinging arms with the elbow bent; on the other hand, in hand-raising motion as the acceleration of arms and following of the entire body.
This is why it is difficult to learn a general representation of the style from limited content motions.
Therefore, we trained the motion for each content in this paper to improve transfer quality from a limited set of motion data.

\subsection{Results}

\begin{figure*}[tb]
  \centering
  \includegraphics[width=0.8\linewidth]{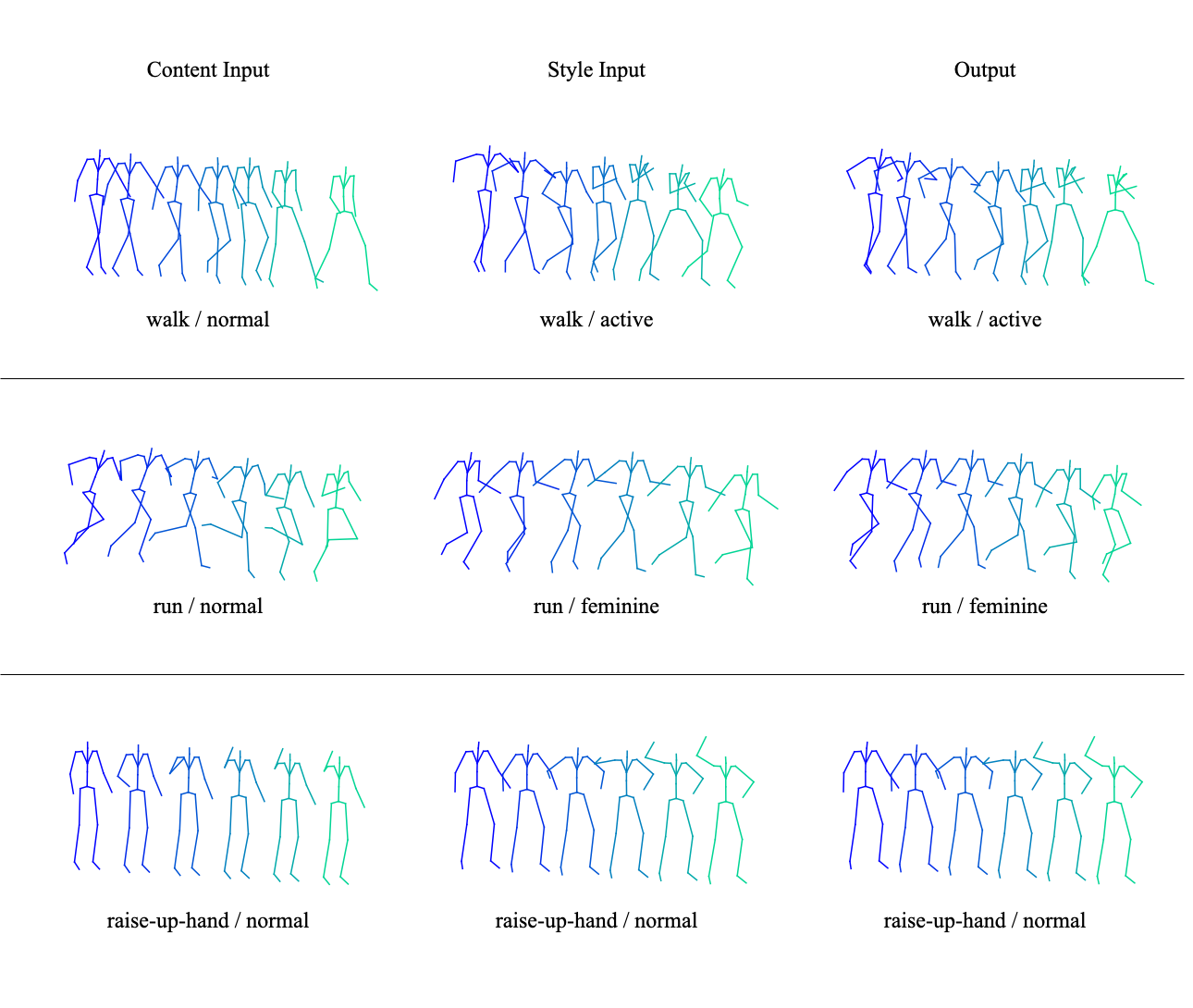}
  \caption{Samples of motion style transfer results using our dataset. Each row corresponds to the transfer set; motion style transfer with the content input (left) and the style input (center) results output (right). Each skeleton is arranged in time sequence from left (blue) to the right (green).}
  \label{fig:results_transfer}
\end{figure*}

Figure \ref{fig:results_transfer} shows the results of motion style transfer using our dataset.
Each row corresponds to the transfer set. 
The model generates output motion (right column) from content input motion (left column) and style input motion (center column).
The motion is successfully transferred, preserving the content of the content input and reflecting the style of the style input.

\begin{figure*}[tb]
  \centering
  \includegraphics[width=\linewidth]{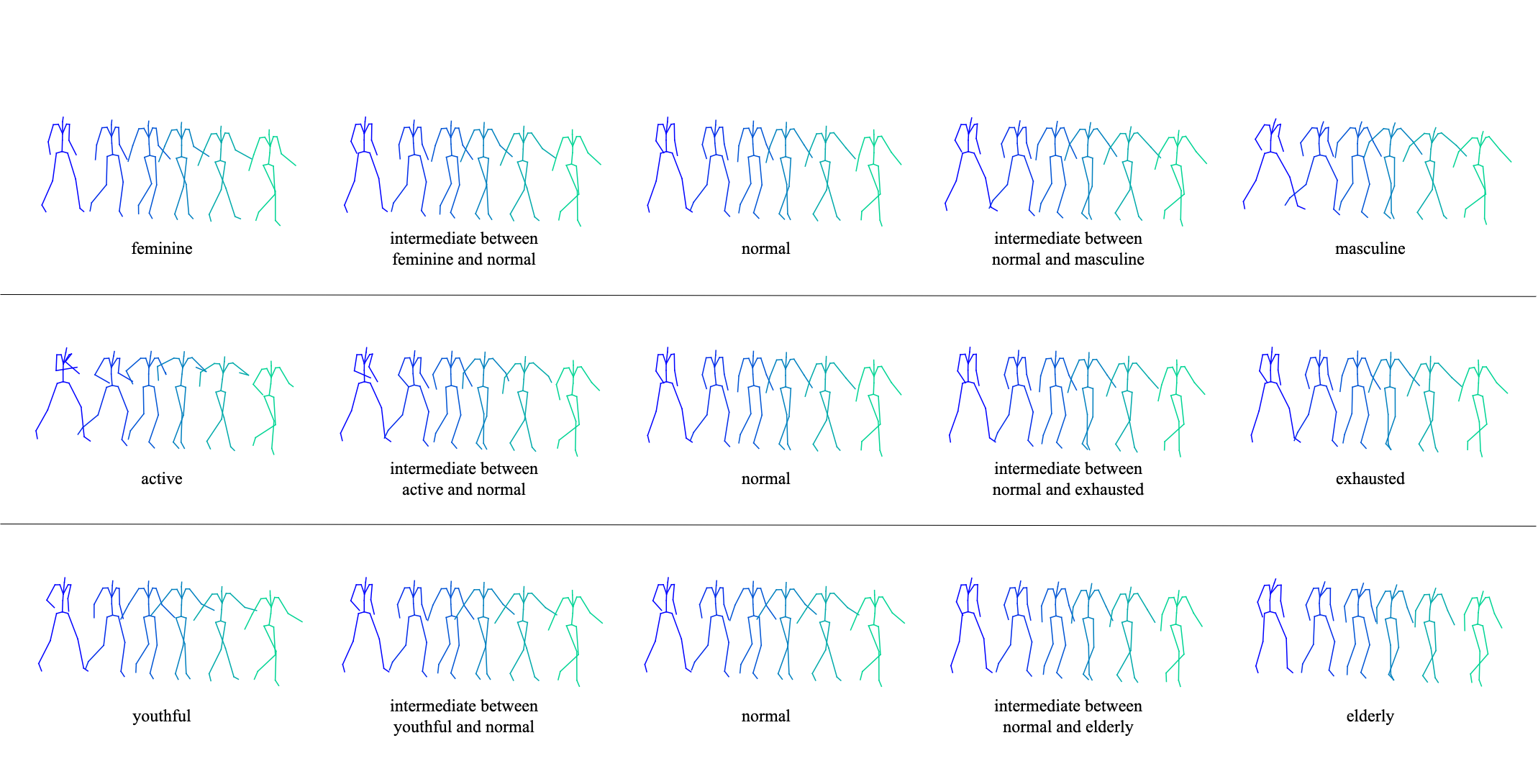}
  \caption{Samples of style interpolation results using our dataset. Each row corresponds to a set of interpolation results of pairwise styles via normal; feminine-normal-masculine (top), active-normal-exhausted (middle), youthful-normal-elderly (bottom). Each skeleton is arranged in time sequence from left (blue) to the right (green).}
  \label{fig:results_interpolation}
\end{figure*}

\begin{figure*}[tb]
  \centering
  \includegraphics[width=0.8\linewidth]{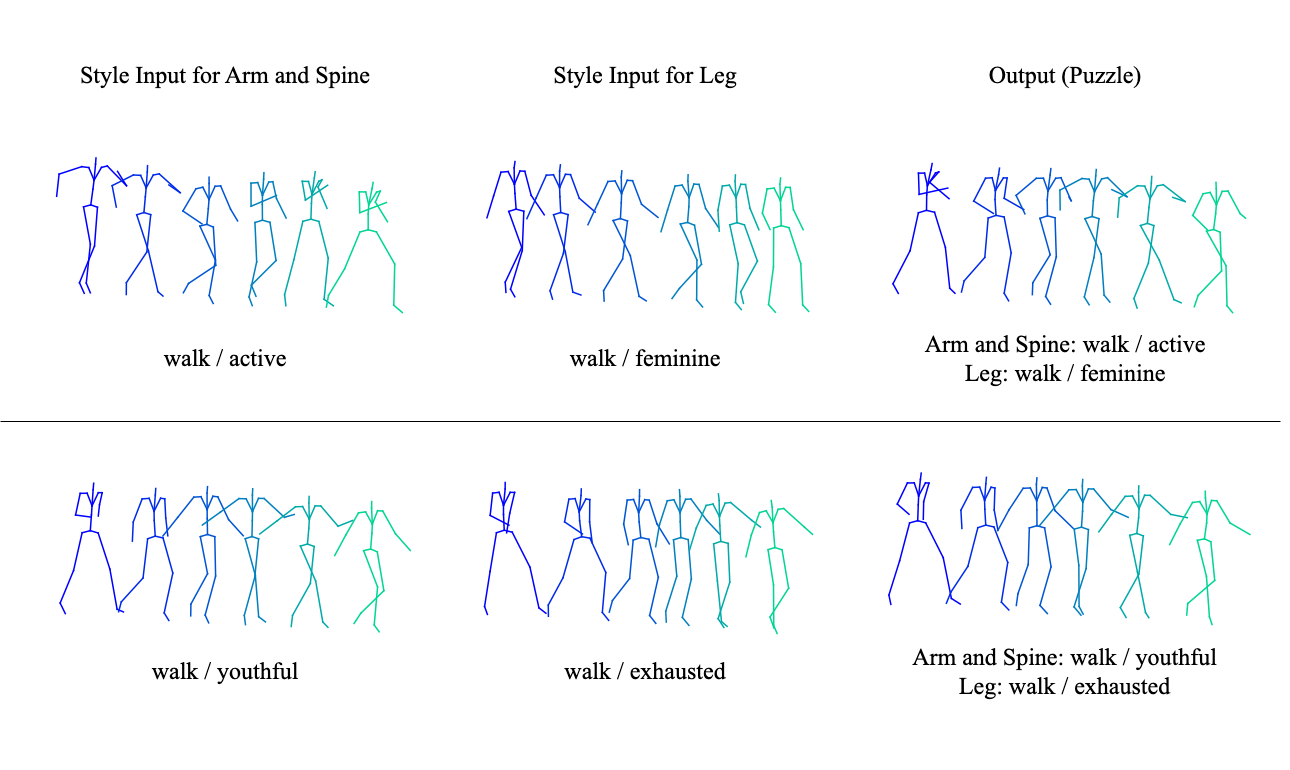}
  \caption{Samples of puzzle results using our dataset. Each row corresponds to a set of puzzle results from two style inputs; puzzle results of active and feminine (top), and puzzle results of youthful and exhausted (bottom). Each skeleton is arranged in time sequence from left (blue) to the right (green).}
  \label{fig:results_puzzle}
\end{figure*}

In addition, we have experimented with the proposed motion dataset using motion interpolation and style transfer on different body parts.
The motion data can be learned to form a smooth latent style space in the embedding layer of the network and can thus generate intermediate motion results that are unseen in the dataset (Figure~\ref{fig:results_interpolation}).
Moreover, taking the superiority of the Motion Puzzle, we are able to produce various stylized motion that has different styles at different body parts (Figure~\ref{fig:results_puzzle}).
Based on these results, we have confirmed the suitability of the dataset for the task of motion style transferring.

\begin{figure*}[tb]
  \centering
  \includegraphics[width=0.8\linewidth]{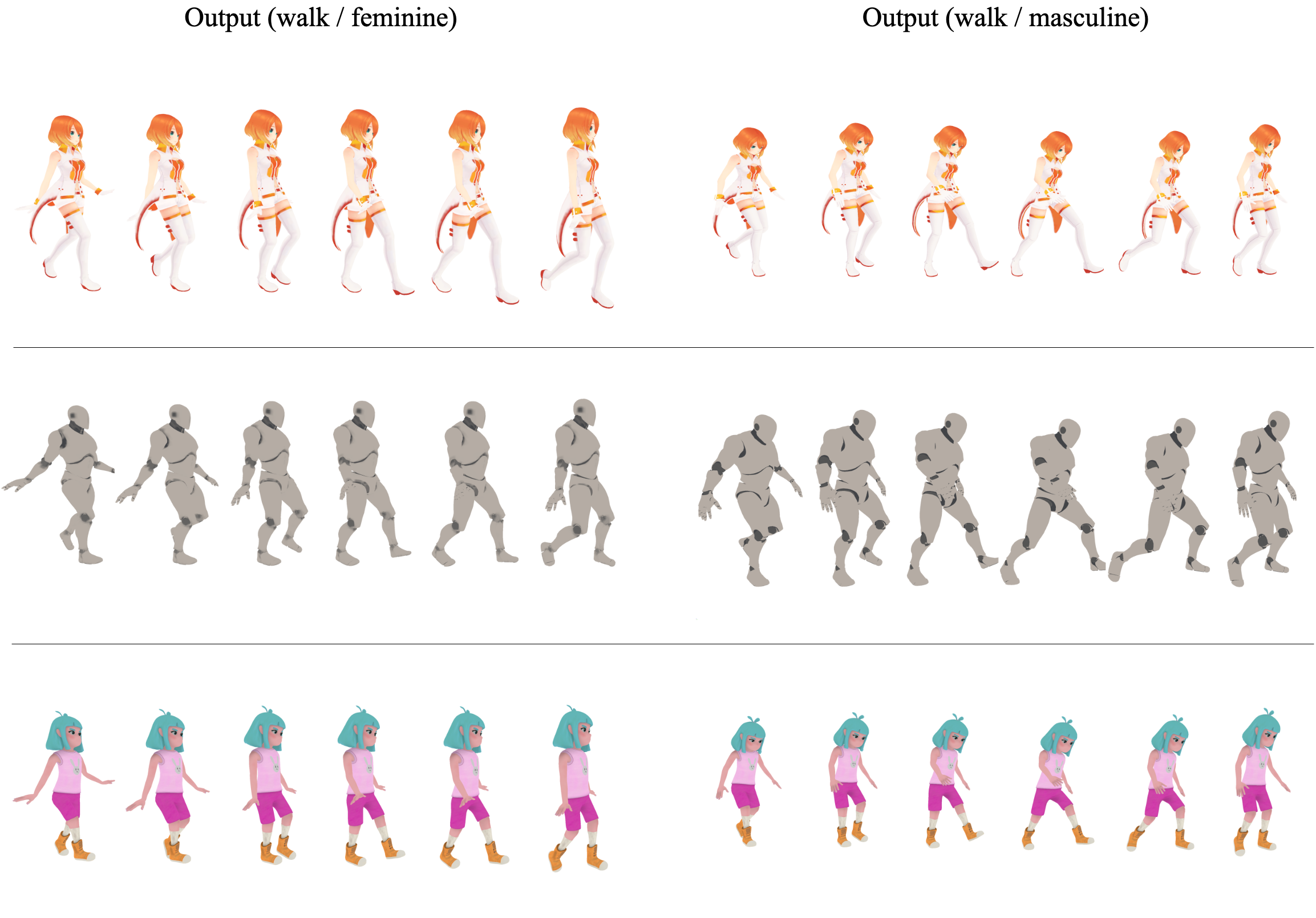}
  \caption{Samples of motion style transfer results using our dataset, applied to various characters. Top: Mirai Komachi. Middle: Unreal Engine Mannequin. Bottom: Mixamo.}
  \label{fig:results_character}
\end{figure*}

Finally, we have performed motion retargeting from the motion data to 3D characters using Unreal Engine 5.
As shown in Figure~\ref{fig:results_character}, the proposed motion data can be easily plugged into both the original character (Mirai Komachi) and other characters from different public databases.

\section{Conclusion}
In this paper, a motion dataset is proposed for the use of motion style transfer.
The stylized motion is captured using professional actors with a large-scale optical motion capture system.
We compare the proposed motion dataset to other datasets used in recent motion style transfer works.
A comprehensive experiment has been done and shows the suitability of the dataset for motion-to-motion transfer.

To contribute to both the research community and the industry, we release the motion dataset to \href{https://github.com/BandaiNamcoResearchInc/Bandai-Namco-Research-Motiondataset}{GitHub} for public usage.
In addition, another motion dataset, which contains advanced motion with more emoting styles, is released to the market for professional usage.
All the motion in the two datasets has an industrial standard human bone structure and is cleaned to be used directly for any application.

In the future, we plan to expand the motion dataset by adding more actors and motion categories.
Moreover, we are building a motion editing pipeline for generating various motions for character control in real applications.
The pipeline includes fixing raw motion data, rigging, and retargeting to the target character for animations and games.
Furthermore, to solve the problem in limited motion data, we urge the development of a universal motion style transfer that learns general representation from the entire motion dataset and thus can transfer style among different motion contents.
We plan to conduct further experiments for the motion style transfer and find the best implementation for industry usage.
When ready, we will embed motion stylization into the motion pipeline and release an industry-ready application.

\bibliographystyle{ACM-Reference-Format}
\bibliography{reference}


\begin{thebibliography}{36}


\ifx \showCODEN    \undefined \def \showCODEN     #1{\unskip}     \fi
\ifx \showDOI      \undefined \def \showDOI       #1{#1}\fi
\ifx \showISBNx    \undefined \def \showISBNx     #1{\unskip}     \fi
\ifx \showISBNxiii \undefined \def \showISBNxiii  #1{\unskip}     \fi
\ifx \showISSN     \undefined \def \showISSN      #1{\unskip}     \fi
\ifx \showLCCN     \undefined \def \showLCCN      #1{\unskip}     \fi
\ifx \shownote     \undefined \def \shownote      #1{#1}          \fi
\ifx \showarticletitle \undefined \def \showarticletitle #1{#1}   \fi
\ifx \showURL      \undefined \def \showURL       {\relax}        \fi
\providecommand\bibfield[2]{#2}
\providecommand\bibinfo[2]{#2}
\providecommand\natexlab[1]{#1}
\providecommand\showeprint[2][]{arXiv:#2}

\bibitem[Aberman et~al\mbox{.}(2020)]%
        {aberman2020unpaired}
\bibfield{author}{\bibinfo{person}{Kfir Aberman}, \bibinfo{person}{Yijia Weng},
  \bibinfo{person}{Dani Lischinski}, \bibinfo{person}{Daniel Cohen-Or}, {and}
  \bibinfo{person}{Baoquan Chen}.} \bibinfo{year}{2020}\natexlab{}.
\newblock \showarticletitle{Unpaired motion style transfer from video to
  animation}.
\newblock \bibinfo{journal}{\emph{ACM Transactions on Graphics (TOG)}}
  \bibinfo{volume}{39}, \bibinfo{number}{4} (\bibinfo{year}{2020}),
  \bibinfo{pages}{64--1}.
\newblock


\bibitem[Akhter and Black(2015)]%
        {akhter2015limitmocap}
\bibfield{author}{\bibinfo{person}{Ijaz Akhter} {and}
  \bibinfo{person}{Michael~J. Black}.} \bibinfo{year}{2015}\natexlab{}.
\newblock \showarticletitle{Pose-conditioned joint angle limits for 3D human
  pose reconstruction}. In \bibinfo{booktitle}{\emph{2015 IEEE Conference on
  Computer Vision and Pattern Recognition (CVPR)}}.
  \bibinfo{pages}{1446--1455}.
\newblock
\urldef\tempurl%
\url{https://doi.org/10.1109/CVPR.2015.7298751}
\showDOI{\tempurl}


\bibitem[Amaya et~al\mbox{.}(1996)]%
        {Amaya1996}
\bibfield{author}{\bibinfo{person}{Kenji Amaya}, \bibinfo{person}{Armin
  Bruderlin}, {and} \bibinfo{person}{Tom Calvert}.}
  \bibinfo{year}{1996}\natexlab{}.
\newblock \showarticletitle{Emotion from Motion}. In
  \bibinfo{booktitle}{\emph{Proceedings of the Graphics Interface 1996
  Conference, May 22-24, 1996, Toronto, Ontario, Canada}}.
  \bibinfo{publisher}{Canadian Human-Computer Communications Society},
  \bibinfo{pages}{222--229}.
\newblock
\showISBNx{0-9695338-5-3}
\urldef\tempurl%
\url{http://graphicsinterface.org/wp-content/uploads/gi1996-26.pdf}
\showURL{%
\tempurl}


\bibitem[BandaiNamcoResearch(2023)]%
        {miraikomachi}
\bibfield{author}{\bibinfo{person}{Inc. BandaiNamcoResearch}.}
  \bibinfo{year}{2023}\natexlab{}.
\newblock \bibinfo{title}{Mirai Komachi Official Page}.
\newblock
\newblock
\urldef\tempurl%
\url{https://www.miraikomachi.com/}
\showURL{%
\tempurl}


\bibitem[Cao et~al\mbox{.}(2021)]%
        {cao2021openpose}
\bibfield{author}{\bibinfo{person}{Zhe Cao}, \bibinfo{person}{Gines Hidalgo},
  \bibinfo{person}{Tomas Simon}, \bibinfo{person}{Shih-En Wei}, {and}
  \bibinfo{person}{Yaser Sheikh}.} \bibinfo{year}{2021}\natexlab{}.
\newblock \showarticletitle{OpenPose: Realtime Multi-Person 2D Pose Estimation
  Using Part Affinity Fields}.
\newblock \bibinfo{journal}{\emph{IEEE Transactions on Pattern Analysis and
  Machine Intelligence}} \bibinfo{volume}{43}, \bibinfo{number}{1}
  (\bibinfo{year}{2021}), \bibinfo{pages}{172--186}.
\newblock
\urldef\tempurl%
\url{https://doi.org/10.1109/TPAMI.2019.2929257}
\showDOI{\tempurl}


\bibitem[Chan et~al\mbox{.}(2011)]%
        {chan2011dance}
\bibfield{author}{\bibinfo{person}{Jacky~C.P. Chan}, \bibinfo{person}{Howard
  Leung}, \bibinfo{person}{Jeff~K.T. Tang}, {and} \bibinfo{person}{Taku
  Komura}.} \bibinfo{year}{2011}\natexlab{}.
\newblock \showarticletitle{A Virtual Reality Dance Training System Using
  Motion Capture Technology}.
\newblock \bibinfo{journal}{\emph{IEEE Transactions on Learning Technologies}}
  \bibinfo{volume}{4}, \bibinfo{number}{2} (\bibinfo{year}{2011}),
  \bibinfo{pages}{187--195}.
\newblock
\urldef\tempurl%
\url{https://doi.org/10.1109/TLT.2010.27}
\showDOI{\tempurl}


\bibitem[CMU(2000)]%
        {CMUMocap}
\bibfield{author}{\bibinfo{person}{Graphics~Lab CMU}.}
  \bibinfo{year}{2000}\natexlab{}.
\newblock \bibinfo{title}{CMU Graphics Lab Motion Capture Database}.
\newblock
\newblock
\urldef\tempurl%
\url{http://mocap.cs.cmu.edu/}
\showURL{%
\tempurl}


\bibitem[Guerra-Filho(2005)]%
        {GuerraFilho2005OpticalMC}
\bibfield{author}{\bibinfo{person}{Gutemberg Guerra-Filho}.}
  \bibinfo{year}{2005}\natexlab{}.
\newblock \showarticletitle{Optical Motion Capture: Theory and Implementation}.
\newblock \bibinfo{journal}{\emph{RITA}}  \bibinfo{volume}{12}
  (\bibinfo{year}{2005}), \bibinfo{pages}{61--90}.
\newblock


\bibitem[Harvey et~al\mbox{.}(2020)]%
        {harvey2020robust}
\bibfield{author}{\bibinfo{person}{Félix~G. Harvey}, \bibinfo{person}{Mike
  Yurick}, \bibinfo{person}{Derek Nowrouzezahrai}, {and}
  \bibinfo{person}{Christopher Pal}.} \bibinfo{year}{2020}\natexlab{}.
\newblock \showarticletitle{Robust Motion In-Betweening}.
\newblock  \bibinfo{volume}{39}, \bibinfo{number}{4} (\bibinfo{year}{2020}).
\newblock


\bibitem[Holden et~al\mbox{.}(2016)]%
        {Holden2016}
\bibfield{author}{\bibinfo{person}{Daniel Holden}, \bibinfo{person}{Jun Saito},
  {and} \bibinfo{person}{Taku Komura}.} \bibinfo{year}{2016}\natexlab{}.
\newblock \showarticletitle{A Deep Learning Framework for Character Motion
  Synthesis and Editing}.
\newblock \bibinfo{journal}{\emph{ACM Trans. Graph.}} \bibinfo{volume}{35},
  \bibinfo{number}{4}, Article \bibinfo{articleno}{138} (\bibinfo{date}{jul}
  \bibinfo{year}{2016}), \bibinfo{numpages}{11}~pages.
\newblock
\showISSN{0730-0301}
\urldef\tempurl%
\url{https://doi.org/10.1145/2897824.2925975}
\showDOI{\tempurl}


\bibitem[Holden et~al\mbox{.}(2015)]%
        {Holden2015}
\bibfield{author}{\bibinfo{person}{Daniel Holden}, \bibinfo{person}{Jun Saito},
  \bibinfo{person}{Taku Komura}, {and} \bibinfo{person}{Thomas Joyce}.}
  \bibinfo{year}{2015}\natexlab{}.
\newblock \showarticletitle{Learning Motion Manifolds with Convolutional
  Autoencoders}. In \bibinfo{booktitle}{\emph{SIGGRAPH Asia 2015 Technical
  Briefs}} (Kobe, Japan) \emph{(\bibinfo{series}{SA '15})}.
  \bibinfo{publisher}{Association for Computing Machinery},
  \bibinfo{address}{New York, NY, USA}, Article \bibinfo{articleno}{18},
  \bibinfo{numpages}{4}~pages.
\newblock
\showISBNx{9781450339308}
\urldef\tempurl%
\url{https://doi.org/10.1145/2820903.2820918}
\showDOI{\tempurl}


\bibitem[Hoyet et~al\mbox{.}(2012)]%
        {hoyet2012sleightmocap}
\bibfield{author}{\bibinfo{person}{Ludovic Hoyet}, \bibinfo{person}{Kenneth
  Ryall}, \bibinfo{person}{Rachel McDonnell}, {and} \bibinfo{person}{Carol
  O'Sullivan}.} \bibinfo{year}{2012}\natexlab{}.
\newblock \showarticletitle{Sleight of Hand: Perception of Finger Motion from
  Reduced Marker Sets}. In \bibinfo{booktitle}{\emph{Proceedings of the ACM
  SIGGRAPH Symposium on Interactive 3D Graphics and Games}} (Costa Mesa,
  California) \emph{(\bibinfo{series}{I3D '12})}.
  \bibinfo{publisher}{Association for Computing Machinery},
  \bibinfo{address}{New York, NY, USA}, \bibinfo{pages}{79–86}.
\newblock
\showISBNx{9781450311946}
\urldef\tempurl%
\url{https://doi.org/10.1145/2159616.2159630}
\showDOI{\tempurl}


\bibitem[Huang and Belongie(2017)]%
        {huang2017arbitrary}
\bibfield{author}{\bibinfo{person}{Xun Huang} {and} \bibinfo{person}{Serge
  Belongie}.} \bibinfo{year}{2017}\natexlab{}.
\newblock \showarticletitle{Arbitrary style transfer in real-time with adaptive
  instance normalization}. In \bibinfo{booktitle}{\emph{Proceedings of the IEEE
  international conference on computer vision}}. \bibinfo{pages}{1501--1510}.
\newblock


\bibitem[Ionescu et~al\mbox{.}(2014)]%
        {Ionescu2014human36}
\bibfield{author}{\bibinfo{person}{Catalin Ionescu}, \bibinfo{person}{Dragos
  Papava}, \bibinfo{person}{Vlad Olaru}, {and} \bibinfo{person}{Cristian
  Sminchisescu}.} \bibinfo{year}{2014}\natexlab{}.
\newblock \showarticletitle{Human3.6M: Large Scale Datasets and Predictive
  Methods for 3D Human Sensing in Natural Environments}.
\newblock \bibinfo{journal}{\emph{IEEE Transactions on Pattern Analysis and
  Machine Intelligence}} \bibinfo{volume}{36}, \bibinfo{number}{7}
  (\bibinfo{date}{jul} \bibinfo{year}{2014}), \bibinfo{pages}{1325--1339}.
\newblock


\bibitem[Jang et~al\mbox{.}(2022)]%
        {jang2022motion}
\bibfield{author}{\bibinfo{person}{Deok-Kyeong Jang}, \bibinfo{person}{Soomin
  Park}, {and} \bibinfo{person}{Sung-Hee Lee}.}
  \bibinfo{year}{2022}\natexlab{}.
\newblock \showarticletitle{Motion Puzzle: Arbitrary Motion Style Transfer by
  Body Part}.
\newblock \bibinfo{journal}{\emph{ACM Transactions on Graphics (TOG)}}
  (\bibinfo{year}{2022}).
\newblock


\bibitem[Liao et~al\mbox{.}(2022)]%
        {Liao2022AIGolf}
\bibfield{author}{\bibinfo{person}{Chen-Chieh Liao}, \bibinfo{person}{Dong-Hyun
  Hwang}, {and} \bibinfo{person}{Hideki Koike}.}
  \bibinfo{year}{2022}\natexlab{}.
\newblock \showarticletitle{AI Golf: Golf Swing Analysis Tool for
  Self-Training}.
\newblock \bibinfo{journal}{\emph{IEEE Access}}  \bibinfo{volume}{10}
  (\bibinfo{year}{2022}), \bibinfo{pages}{106286--106295}.
\newblock
\urldef\tempurl%
\url{https://doi.org/10.1109/ACCESS.2022.3210261}
\showDOI{\tempurl}


\bibitem[Liao et~al\mbox{.}(2023)]%
        {Liao2023AICoach}
\bibfield{author}{\bibinfo{person}{Chen-Chieh Liao}, \bibinfo{person}{Dong-Hyun
  Hwang}, \bibinfo{person}{Erwin Wu}, {and} \bibinfo{person}{Hideki Koike}.}
  \bibinfo{year}{2023}\natexlab{}.
\newblock \showarticletitle{AI Coach: A Motor Skill Training System Using
  Motion Discrepancy Detection}. In \bibinfo{booktitle}{\emph{Proceedings of
  the Augmented Humans International Conference 2023}} (Glasgow, United
  Kingdom) \emph{(\bibinfo{series}{AHs '23})}. \bibinfo{publisher}{Association
  for Computing Machinery}, \bibinfo{address}{New York, NY, USA},
  \bibinfo{pages}{179–189}.
\newblock
\showISBNx{9781450399845}
\urldef\tempurl%
\url{https://doi.org/10.1145/3582700.3582710}
\showDOI{\tempurl}


\bibitem[Ma et~al\mbox{.}(2010)]%
        {Ma2010}
\bibfield{author}{\bibinfo{person}{Wanli Ma}, \bibinfo{person}{Shihong Xia},
  \bibinfo{person}{Jessica~K. Hodgins}, \bibinfo{person}{Xiao Yang},
  \bibinfo{person}{Chunpeng Li}, {and} \bibinfo{person}{Zhaoqi Wang}.}
  \bibinfo{year}{2010}\natexlab{}.
\newblock \showarticletitle{{Modeling Style and Variation in Human Motion}}. In
  \bibinfo{booktitle}{\emph{Eurographics/ ACM SIGGRAPH Symposium on Computer
  Animation}}, \bibfield{editor}{\bibinfo{person}{MZoran Popovic} {and}
  \bibinfo{person}{Miguel Otaduy}} (Eds.). \bibinfo{publisher}{The Eurographics
  Association}.
\newblock
\showISBNx{978-3-905674-27-9}
\showISSN{1727-5288}
\urldef\tempurl%
\url{https://doi.org/10.2312/SCA/SCA10/021-030}
\showDOI{\tempurl}


\bibitem[Mandery et~al\mbox{.}(2015)]%
        {mandery2015KITmocap}
\bibfield{author}{\bibinfo{person}{Christian Mandery}, \bibinfo{person}{Ömer
  Terlemez}, \bibinfo{person}{Martin Do}, \bibinfo{person}{Nikolaus
  Vahrenkamp}, {and} \bibinfo{person}{Tamim Asfour}.}
  \bibinfo{year}{2015}\natexlab{}.
\newblock \showarticletitle{The KIT whole-body human motion database}. In
  \bibinfo{booktitle}{\emph{2015 International Conference on Advanced Robotics
  (ICAR)}}. \bibinfo{pages}{329--336}.
\newblock
\urldef\tempurl%
\url{https://doi.org/10.1109/ICAR.2015.7251476}
\showDOI{\tempurl}


\bibitem[Mason et~al\mbox{.}(2022)]%
        {mason2022realtime}
\bibfield{author}{\bibinfo{person}{Ian Mason}, \bibinfo{person}{Sebastian
  Starke}, {and} \bibinfo{person}{Taku Komura}.}
  \bibinfo{year}{2022}\natexlab{}.
\newblock \showarticletitle{Real-Time Style Modelling of Human Locomotion via
  Feature-Wise Transformations and Local Motion Phases}.
\newblock \bibinfo{journal}{\emph{arXiv preprint arXiv:2201.04439}}
  (\bibinfo{year}{2022}).
\newblock


\bibitem[Moryossef et~al\mbox{.}(2020)]%
        {moryossef2020sign}
\bibfield{author}{\bibinfo{person}{Amit Moryossef}, \bibinfo{person}{Ioannis
  Tsochantaridis}, \bibinfo{person}{Roee Aharoni}, \bibinfo{person}{Sarah
  Ebling}, {and} \bibinfo{person}{Srini Narayanan}.}
  \bibinfo{year}{2020}\natexlab{}.
\newblock \showarticletitle{Real-Time Sign Language Detection Using Human Pose
  Estimation}. In \bibinfo{booktitle}{\emph{Computer Vision -- ECCV 2020
  Workshops}}, \bibfield{editor}{\bibinfo{person}{Adrien Bartoli} {and}
  \bibinfo{person}{Andrea Fusiello}} (Eds.). \bibinfo{publisher}{Springer
  International Publishing}, \bibinfo{address}{Cham},
  \bibinfo{pages}{237--248}.
\newblock
\showISBNx{978-3-030-66096-3}


\bibitem[Movella(nd)]%
        {Xsens}
\bibfield{author}{\bibinfo{person}{Movella}.} \bibinfo{year}{n.d}\natexlab{}.
\newblock \bibinfo{title}{Xsens MVN MoCap Systems}.
\newblock
\newblock
\urldef\tempurl%
\url{https://www.vicon.com/}
\showURL{%
\tempurl}


\bibitem[M\"{u}ller et~al\mbox{.}(2009)]%
        {muller2009efficientmocap}
\bibfield{author}{\bibinfo{person}{Meinard M\"{u}ller},
  \bibinfo{person}{Andreas Baak}, {and} \bibinfo{person}{Hans-Peter Seidel}.}
  \bibinfo{year}{2009}\natexlab{}.
\newblock \showarticletitle{Efficient and Robust Annotation of Motion Capture
  Data}. In \bibinfo{booktitle}{\emph{Proceedings of the 2009 ACM
  SIGGRAPH/Eurographics Symposium on Computer Animation}} (New Orleans,
  Louisiana) \emph{(\bibinfo{series}{SCA '09})}.
  \bibinfo{publisher}{Association for Computing Machinery},
  \bibinfo{address}{New York, NY, USA}, \bibinfo{pages}{17–26}.
\newblock
\showISBNx{9781605586106}
\urldef\tempurl%
\url{https://doi.org/10.1145/1599470.1599473}
\showDOI{\tempurl}


\bibitem[Park et~al\mbox{.}(2021)]%
        {park2021diverse}
\bibfield{author}{\bibinfo{person}{Soomin Park}, \bibinfo{person}{Deok-Kyeong
  Jang}, {and} \bibinfo{person}{Sung-Hee Lee}.}
  \bibinfo{year}{2021}\natexlab{}.
\newblock \showarticletitle{Diverse Motion Stylization for Multiple Style
  Domains via Spatial-Temporal Graph-Based Generative Model}.
\newblock \bibinfo{journal}{\emph{Proceedings of the ACM on Computer Graphics
  and Interactive Techniques}} \bibinfo{volume}{4}, \bibinfo{number}{3}
  (\bibinfo{year}{2021}), \bibinfo{pages}{1--17}.
\newblock


\bibitem[Rincon et~al\mbox{.}(2016)]%
        {rincon2016gamemocap}
\bibfield{author}{\bibinfo{person}{Alejandro~Lopez Rincon},
  \bibinfo{person}{Hiroshi Yamasaki}, {and} \bibinfo{person}{Shingo Shimoda}.}
  \bibinfo{year}{2016}\natexlab{}.
\newblock \showarticletitle{Design of a video game for rehabilitation using
  motion capture, EMG analysis and virtual reality}. In
  \bibinfo{booktitle}{\emph{2016 International Conference on Electronics,
  Communications and Computers (CONIELECOMP)}}. \bibinfo{pages}{198--204}.
\newblock
\urldef\tempurl%
\url{https://doi.org/10.1109/CONIELECOMP.2016.7438575}
\showDOI{\tempurl}


\bibitem[Roetenberg et~al\mbox{.}(2009)]%
        {Roetenberg2009Xsens}
\bibfield{author}{\bibinfo{person}{Daniel Roetenberg}, \bibinfo{person}{Henk
  Luinge}, {and} \bibinfo{person}{Per Slycke}.}
  \bibinfo{year}{2009}\natexlab{}.
\newblock \showarticletitle{Xsens MVN: Full 6DOF human motion tracking using
  miniature inertial sensors}.
\newblock \bibinfo{journal}{\emph{Xsens Motion Technol. BV Tech. Rep.}}
  \bibinfo{volume}{3} (\bibinfo{date}{01} \bibinfo{year}{2009}).
\newblock


\bibitem[Shiro et~al\mbox{.}(2019)]%
        {shiro2019interpose}
\bibfield{author}{\bibinfo{person}{Keisuke Shiro}, \bibinfo{person}{Kazme
  Egawa}, \bibinfo{person}{Takashi Miyaki}, {and} \bibinfo{person}{Jun
  Rekimoto}.} \bibinfo{year}{2019}\natexlab{}.
\newblock \showarticletitle{InterPoser: Visualizing Interpolated Movements for
  Bouldering Training}. In \bibinfo{booktitle}{\emph{Extended Abstracts of the
  2019 CHI Conference on Human Factors in Computing Systems}} (Glasgow,
  Scotland Uk) \emph{(\bibinfo{series}{CHI EA '19})}.
  \bibinfo{publisher}{Association for Computing Machinery},
  \bibinfo{address}{New York, NY, USA}, \bibinfo{pages}{1–6}.
\newblock
\showISBNx{9781450359719}
\urldef\tempurl%
\url{https://doi.org/10.1145/3290607.3312779}
\showDOI{\tempurl}


\bibitem[Troje(2002)]%
        {Troje2002DecomposingBM}
\bibfield{author}{\bibinfo{person}{Nikolaus~F. Troje}.}
  \bibinfo{year}{2002}\natexlab{}.
\newblock \showarticletitle{Decomposing biological motion: a framework for
  analysis and synthesis of human gait patterns.}
\newblock \bibinfo{journal}{\emph{Journal of vision}}  \bibinfo{volume}{2 5}
  (\bibinfo{year}{2002}), \bibinfo{pages}{371--87}.
\newblock


\bibitem[Unuma et~al\mbox{.}(1995)]%
        {Unuma1995}
\bibfield{author}{\bibinfo{person}{Munetoshi Unuma}, \bibinfo{person}{Ken
  Anjyo}, {and} \bibinfo{person}{Ryozo Takeuchi}.}
  \bibinfo{year}{1995}\natexlab{}.
\newblock \showarticletitle{Fourier Principles for Emotion-Based Human Figure
  Animation}. In \bibinfo{booktitle}{\emph{Proceedings of the 22nd Annual
  Conference on Computer Graphics and Interactive Techniques}}
  \emph{(\bibinfo{series}{SIGGRAPH '95})}. \bibinfo{publisher}{Association for
  Computing Machinery}, \bibinfo{address}{New York, NY, USA},
  \bibinfo{pages}{91–96}.
\newblock
\showISBNx{0897917014}
\urldef\tempurl%
\url{https://doi.org/10.1145/218380.218419}
\showDOI{\tempurl}


\bibitem[Vicon(nd)]%
        {Vicon}
\bibfield{author}{\bibinfo{person}{Vicon}.} \bibinfo{year}{n.d}\natexlab{}.
\newblock \bibinfo{title}{Vicon Motion Systems}.
\newblock
\newblock
\urldef\tempurl%
\url{https://www.vicon.com/}
\showURL{%
\tempurl}


\bibitem[Wang et~al\mbox{.}(2021a)]%
        {Huaijun2021}
\bibfield{author}{\bibinfo{person}{Huaijun Wang}, \bibinfo{person}{Dandan Du},
  \bibinfo{person}{Junhuai li}, \bibinfo{person}{Wenchao Ji}, {and}
  \bibinfo{person}{Lei Yu}.} \bibinfo{year}{2021}\natexlab{a}.
\newblock \showarticletitle{A Cyclic Consistency Motion Style Transfer Method
  Combined with Kinematic Constraints}.
\newblock \bibinfo{journal}{\emph{Journal of Sensors}}  \bibinfo{volume}{2021}
  (\bibinfo{date}{06} \bibinfo{year}{2021}), \bibinfo{pages}{1--17}.
\newblock
\urldef\tempurl%
\url{https://doi.org/10.1155/2021/5548614}
\showDOI{\tempurl}


\bibitem[Wang et~al\mbox{.}(2021b)]%
        {wang2021hrnet}
\bibfield{author}{\bibinfo{person}{Jingdong Wang}, \bibinfo{person}{Ke Sun},
  \bibinfo{person}{Tianheng Cheng}, \bibinfo{person}{Borui Jiang},
  \bibinfo{person}{Chaorui Deng}, \bibinfo{person}{Yang Zhao},
  \bibinfo{person}{Dong Liu}, \bibinfo{person}{Yadong Mu},
  \bibinfo{person}{Mingkui Tan}, \bibinfo{person}{Xinggang Wang},
  \bibinfo{person}{Wenyu Liu}, {and} \bibinfo{person}{Bin Xiao}.}
  \bibinfo{year}{2021}\natexlab{b}.
\newblock \showarticletitle{Deep High-Resolution Representation Learning for
  Visual Recognition}.
\newblock \bibinfo{journal}{\emph{IEEE Transactions on Pattern Analysis and
  Machine Intelligence}} \bibinfo{volume}{43}, \bibinfo{number}{10}
  (\bibinfo{year}{2021}), \bibinfo{pages}{3349--3364}.
\newblock
\urldef\tempurl%
\url{https://doi.org/10.1109/TPAMI.2020.2983686}
\showDOI{\tempurl}


\bibitem[Wen et~al\mbox{.}(2021)]%
        {Yu-Hui2021}
\bibfield{author}{\bibinfo{person}{Yu-Hui Wen}, \bibinfo{person}{Zhipeng Yang},
  \bibinfo{person}{Hongbo Fu}, \bibinfo{person}{Lin Gao},
  \bibinfo{person}{Yanan Sun}, {and} \bibinfo{person}{Yong-Jin Liu}.}
  \bibinfo{year}{2021}\natexlab{}.
\newblock \showarticletitle{Autoregressive Stylized Motion Synthesis with
  Generative Flow}. In \bibinfo{booktitle}{\emph{2021 IEEE/CVF Conference on
  Computer Vision and Pattern Recognition (CVPR)}}.
  \bibinfo{pages}{13607--13607}.
\newblock
\urldef\tempurl%
\url{https://doi.org/10.1109/CVPR46437.2021.01340}
\showDOI{\tempurl}


\bibitem[Xia et~al\mbox{.}(2015)]%
        {Xia2015dataset}
\bibfield{author}{\bibinfo{person}{Shihong Xia}, \bibinfo{person}{Congyi Wang},
  \bibinfo{person}{Jinxiang Chai}, {and} \bibinfo{person}{Jessica Hodgins}.}
  \bibinfo{year}{2015}\natexlab{}.
\newblock \showarticletitle{Realtime Style Transfer for Unlabeled Heterogeneous
  Human Motion}.
\newblock \bibinfo{journal}{\emph{ACM Trans. Graph.}} \bibinfo{volume}{34},
  \bibinfo{number}{4} (\bibinfo{date}{jul} \bibinfo{year}{2015}).
\newblock
\urldef\tempurl%
\url{https://doi.org/10.1145/2766999}
\showDOI{\tempurl}


\bibitem[Yumer and Mitra(2016)]%
        {yumer2016spectral}
\bibfield{author}{\bibinfo{person}{M~Ersin Yumer} {and}
  \bibinfo{person}{Niloy~J Mitra}.} \bibinfo{year}{2016}\natexlab{}.
\newblock \showarticletitle{Spectral style transfer for human motion between
  independent actions}.
\newblock \bibinfo{journal}{\emph{ACM Transactions on Graphics (TOG)}}
  \bibinfo{volume}{35}, \bibinfo{number}{4} (\bibinfo{year}{2016}),
  \bibinfo{pages}{1--8}.
\newblock


\bibitem[Zhu et~al\mbox{.}(2022)]%
        {Zhu2022musclerehab}
\bibfield{author}{\bibinfo{person}{Junyi Zhu}, \bibinfo{person}{Yuxuan Lei},
  \bibinfo{person}{Aashini Shah}, \bibinfo{person}{Gila Schein},
  \bibinfo{person}{Hamid Ghaednia}, \bibinfo{person}{Joseph Schwab},
  \bibinfo{person}{Casper Harteveld}, {and} \bibinfo{person}{Stefanie
  Mueller}.} \bibinfo{year}{2022}\natexlab{}.
\newblock \showarticletitle{MuscleRehab: Improving Unsupervised Physical
  Rehabilitation by Monitoring and Visualizing Muscle Engagement}. In
  \bibinfo{booktitle}{\emph{Proceedings of the 35th Annual ACM Symposium on
  User Interface Software and Technology}} (Bend, OR, USA)
  \emph{(\bibinfo{series}{UIST '22})}. \bibinfo{publisher}{Association for
  Computing Machinery}, \bibinfo{address}{New York, NY, USA}, Article
  \bibinfo{articleno}{33}, \bibinfo{numpages}{14}~pages.
\newblock
\showISBNx{9781450393201}
\urldef\tempurl%
\url{https://doi.org/10.1145/3526113.3545705}
\showDOI{\tempurl}


\end{thebibliography}

\appendix
\section{Appendix}

\subsection{Emote motion dataset}
\label{appendix:emote}

We have collected another dataset, Emote motion dataset that includes complex motions for more practical motion-style transfer.
Different from the two public datasets, motion data in the Emot motion dataset is collected using the Xsens MVN MoCap Systems~\cite{Xsens}. The motion data is sampled with a framerate of 120 FPS.

The dataset consists of a total of 42 types of motions, combining 6 types of content that represent dance motions and 7 types of styles that reflect characterization such as "cute" and "rough". The data set contains a total of 1,475,826 frames in 120 FPS of data that can be used for emotes, a function that lets characters gesture, and dance to express greetings and emotions.

Emote motion dataset is commercially available and is targeted for R\&D, universities, and research institutes that are conducting or planning to conduct Motion AI research.
More information and the inquiry form can be found on the following websites:
\begin{itemize}
    \item English: TBA
    \item Japanese: \href{https://acesinc.co.jp/projects/project-1760}{https://acesinc.co.jp/projects/project-1760}
\end{itemize}

\subsection{Implementation Modification}
\label{appendix:code_modification}
As shown in \ref{fig:diff_born}, the bone structure of our dataset is different from the one of CMU dataset which is used in Motion Puzzle~\cite{jang2022motion}.
Therefore, we change the original implementation to apply it to our motion dataset.

In particular, the number of joints used in the original implementation stays the same as our skeleton.
Therefore, we pick all the joints of our skeleton and match them to the original joints following the order in the hierarchical bone structure.
Table \ref{tab:correspondence_of_joints} shows the joint correspondence of the two bone structures.

In addition, since the shoulder bone is ignored in the original implementation, we adjust the pooling range on hands during the convolution process in the network.
While the pooling around the shoulder is done using the chest, upper arm, and forearm in the original implementation, we have changed to using the shoulder, upper arm, and forearm to apply the average pooling.
In the qualitative evaluation, we have found that the change of pooling targets helps improve the style transfer results. 

\begin{table}[H]
    \caption{Correspondence relation of joints}
    \label{tab:correspondence_of_joints}
    \centering
    \begin{tabular}{rlrl}
        \hline
        Ours order & Ours name & CMU order & CMU name \\
        \hline \hline
        0 & Hips & 0 & Hips \\
        1 & Spine & 12 & Spine \\
        2 & Chest & 13 & Spine1 \\
        3 & Neck & 15 & Neck1 \\
        4 & Head & 16 & Head \\
        5 & Shoulder\_L & 18 & LeftArm \\
        6 & UpperArm\_L & 19 & LeftForeArm \\
        7 & LowerArm\_L & 20 &  LeftHand \\
        8 & Hand\_L & 22 & LeftHandIndex \\
        9 & Shoulder\_R & 25 & RightArm \\
        10 & UpperArm\_R & 26 & RightForeArm \\
        11 & LowerArm\_R & 27 & RightHand \\
        12 & Hand\_R & 29 & RightHandIndex \\
        13 & UpperLeg\_L & 2 & LeftUpLeg \\
        14 & LowerLeg\_L & 3 & LeftLeg \\
        15 & Foot\_L & 4 & LeftFoot \\
        16 & Toe\_L & 5 & LeftToeBase \\
        17 & UpperLeg\_R & 7 & RightUpLeg \\
        18 & LowerLeg\_R & 8 & RightLeg \\
        19 & Foot\_R & 9 & RightFoot \\
        20 & Toe\_R & 10 & RightToeBase \\
        \hline
    \end{tabular}
\end{table}

\end{document}